\documentclass[letterpaper, 10 pt, conference]{ieeeconf} 
\IEEEoverridecommandlockouts             

\usepackage{graphics} 
\usepackage{epsfig} 
\usepackage{times} 
\usepackage{amsmath} 
\usepackage{amssymb}  
\usepackage[per-mode=symbol]{siunitx}

\usepackage[isbn=false,
    backend=biber,
    style=ieee,
    bibstyle=ieee,  
    sortcites=true,   
    giveninits=true,            
    backref=false]{biblatex}
\title{\LARGE \bf
Real-time Optimal Landing Control of the MIT Mini Cheetah
}

\author{Se Hwan Jeon$^{1}$, Sangbae Kim$^{1}$, and Donghyun Kim$^{2}$
\thanks{$^{1}$Se Hwan Jeon and Sangbae Kim are with the Department of Mechanical Engineering,
        Massachusetts Institute of Technology, Cambridge, MA 02139, USA
        {\tt\small sehwan@mit.edu, sangbae@mit.edu}}%
\thanks{$^{2}$Donghyun Kim is with the Department of Mechanical Engineering, University of Massachusetts Amherst,
        Amherst, MA 01003, USA
        {\tt\small donghyunkim@cs.umass.edu}}%
}

\AtBeginBibliography{\footnotesize}

\begin{document}

\maketitle
\thispagestyle{empty}
\pagestyle{empty}

\allowdisplaybreaks
\begin{abstract}
Quadrupedal landing is a complex process involving large impacts, elaborate contact transitions, and is a crucial recovery behavior observed in many biological animals. This work presents a real-time, optimal landing controller that is free of pre-specified contact schedules. The controller determines optimal touchdown postures and reaction force profiles and is able to recover from a variety of falling configurations. The quadrupedal platform used, the MIT Mini Cheetah, recovered safely from drops of up to 8 \si{\meter} in simulation, as well as from a range of orientations and planar velocities. The controller is also tested on hardware, successfully recovering from drops of up to 2 \si{\meter}.  



\end{abstract}




\section{INTRODUCTION}


 Safe recovery from planned drops or unexpected falls is one of the most crucial features of animals and is beneficial both in navigating challenging terrain and preventing significant damage in the case of unexpected drops. Furthermore, a robust landing controller opens up the possibility of deploying quadruped robots directly into harsh environments where supervision can be difficult. However, in contrast to the progress in quadrupedal locomotion on robotic platforms \cite{mc-vision, hutter-RL-locomotion}, there has been relatively little work exploring how robots address significant changes in elevation or react to falling. Several works have explored controlling a body's orientation while rotating, but do not focus on the reaction force profiles and poses needed to safely recover from the fall \cite{Bingham-orient, Kurtz-fallingCat}.
 
\begin{figure}
    \centering
    \includegraphics[width=0.95\columnwidth]{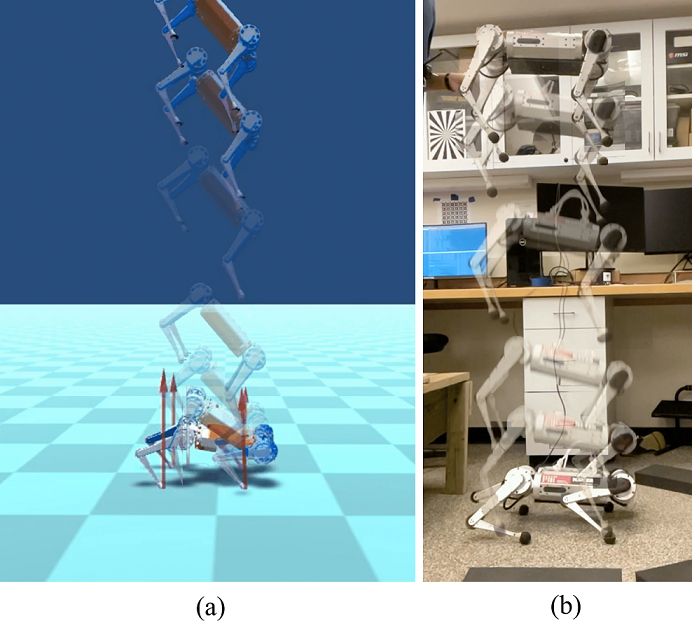}
    \caption{ {\bf High speed landing of quadruped robot.} The proposed controller enables the robot to safely land after falling from a high place. (a) In the simulation, we accomplished safe landing from 2.5\si{\meter}-high free drop. (b) In hardware experiments, we demonstrated 2\si{\meter}-high free drop.}
    \label{fig:mc-landing}
    \vspace{-0.3cm}
\end{figure}

A common approach for landing actuated systems is to control contact point impedances with some additional feedforward reaction forces. Lynch et al. describes the so-called `soft landing problem', exploring control methods to minimize foot penetration depth into a surface from a fall with prismatic actuation, but discussion was limited to single impacts on a body without rigid contacts \cite{Lynch-softLanding}.
Similar to quadrupeds, \cite{Kiefer-rotorcraft} explores actuated landing gear for rotorcraft and finding optimal impedances to mitigate impacts, but considers only horizontal landings. Although previous works demonstrated the effectiveness of impedance control, they did not capture all detailed motion included in landing behavior such as landing posture, reaction force profile, contact location, and sequence. Without taking into account these complexities, we cannot fully utilize the robot hardware's capability to achieve smooth and safe landings. 
\par


For motion planning problems consisting of multiple, complex decision processes, trajectory optimization methods have been widely used. It is worth noting that automatic contact sequence selection is crucial to controlling an landing because feasible contact sequences could be entirely different depending on the orientation of the body at touchdown. This timing becomes especially important at higher velocities. Previous work by Winkler et al. parameterizes the gait to optimize for the timings of the contact schedule, but the complexity of the formulation makes it difficult to solve for real-time applications \cite{Winkler-gaitPhase}. Similarly, the work of \cite{Mordatch-CIO} shows realistic, dynamic behaviors independent of prescribed contacts, but is likewise intractable for real-time performance. 
\par

Several other approaches such as reinforcement learning or model-predictive control (MPC) present interesting landing behaviors also demonstrated on hardware. \cite{hutter-RL-landing} demonstrated planar landing and airborne orientation control on the SpaceBok quadruped in a "low-gravity" environment, but it is unclear how easily the algorithm could be applied for real-world, 3D conditions with more dramatic impacts and inertial effects. While planar landing and jumping was also demonstrated in \cite{quan-mitCheetah3-jumping} and \cite{hutter-starlETH}, touchdown was made without considering optimal touchdown positions or timings, and were from relatively low heights with little pitch or roll of the body. 
With a heuristics-guided MPC, \cite{Bledt-RPC} demonstrated dropping the MIT Mini Cheetah from around 1.5 \si{\meter}, but did not investigate landing with changes in orientation or body velocities. Simple joint or Cartesian impedance control can often be enough to stabilize a robot's motion from small drops, but fail consistently at higher velocities or with significant orientations away from the horizontal. 
\par

In this work, we use a nonlinear trajectory optimization including contact complementary constraints to find optimal landing postures and reaction force profiles. The formulation is based on \cite{Posa-CCC}, and we modified the algorithm to accommodate actuator torque and leg kinematic limits. The optimization is solved while the robot is airborne in a MPC fashion at approximately 10 \si{\hertz}.
Once touchdown is detected, the optimized trajectory is then tracked with a whole-body impulse controller \cite{DHKim-highlyDynamic}.
In our high-fidelity physics simulation environment, the Mini Cheetah robot recovered from drops of up to 8 \si{\meter} high, angular velocities between -0.5 and 0.5 \si{\radian\per\second}, horizontal velocities ranging from -1.5 to 1.5 \si{\meter\per\second}, pitch orientations ranging from $-{\frac{\pi}{3}}$ to $\frac{\pi}{3}$, and roll orientations ranging from $-{\frac{\pi}{6}}$ to $\frac{\pi}{6}$.
In hardware tests, the Mini Cheetah was able to land successfully from a height of 2 \si{\meter}.

The contribution of this work is two-fold: 1) we formulate a real-time landing control framework that finds optimal contact locations and timings from various orientations and body velocities, and 2) we demonstrate successful landings in simulation from significant heights and orientations and in hardware from heights of up to 2 \si{\meter}, as shown in Figure \ref{fig:mc-landing}.

\section{SYSTEM OVERVIEW}

The MIT Mini Cheetah has a mass of approximately 9 kg with 12 modular actuators (ab/ad, hip, and knee for each of its four legs). Each actuator is capable of producing a maximum torque of 17 \si{\newton\meter} and a continuous torque of 6.9 \si{\newton\meter} \cite{Katz-MC}. Because the legs make up less than 10\% of its mass, the dynamics of the Mini Cheetah can be approximated as a single rigid body model (SRBM) given as
\begin{eqnarray} \label{eqn:srbm}
    m \ddot{\mathbf{p}} &=& \sum_{i = 1}^{n_c}\boldsymbol{\lambda}_i - \mathbf{f}_g,\\
    {d \over dt}(\mathbf{I}\boldsymbol{\omega}) &=& \sum_{i = 1}^{n_c}(\mathbf{r}_i-\mathbf{p}) \times \boldsymbol{\lambda}_i,
\end{eqnarray}
where $m$ is the mass, $n_c$ is the number of contacts, $\mathbf{p}$, $\mathbf{r}_i$, $\boldsymbol{\lambda}_i$, and $\mathbf{f}_g$ are the vectors for the body position, foot positions, reaction forces, and gravitational force, expressed in the world frame.  $\mathbf{I} \in \mathbb{R}^{3 \times 3}$ and $\mathbf{\omega}$ are the rotational inertia tensor of the body and angular velocity of the body respectively, expressed in the body frame. 



\section{FRAMEWORK}
\begin{figure} 
    \centering
    \includegraphics[width=1.0\columnwidth]{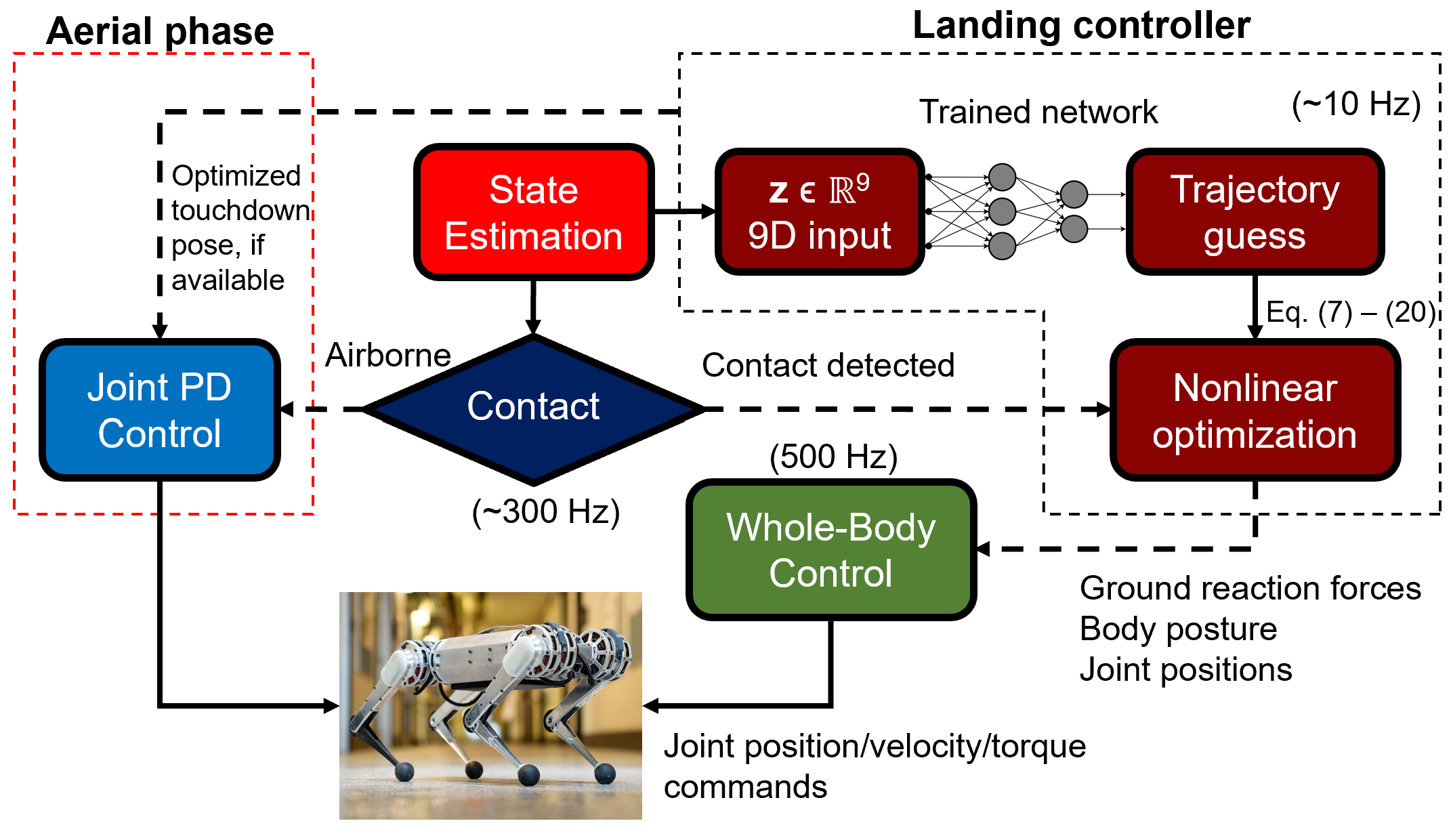}
    \caption{ {\bf Overview of the components of the landing controller.} If contact is detected, the latest optimization solution is tracked with whole-body control. Otherwise, the robot maintains a nominal or optimized pose with joint PD control while airborne.}
    \label{fig:control-diagram}
\end{figure}

The landing controller in this work consists of three major components: nonlinear model predictive control, a trained neural network that outputs initial guesses for trajectories, and a whole-body controller.
Based on its current state estimate, a nonlinear program (NLP) solves for dynamically feasible landing trajectories. The optimization is "warm started" with a guess for the trajectories output by the trained neural network to improve convergence speed.
Finally, when touchdown is detected, the optimized trajectory is used as a reference and tracked by a whole-body controller. An overview of this framework is shown in Figure \ref{fig:control-diagram}.

\subsection{Optimization Formulation}
The controller is formulated as a direct transcription, kino-dynamic trajectory optimization over $N$ timesteps, similar to \cite{Dai-KDTO} and \cite{Chignoli-MITHumanoid} with three key modifications for real-time performance: the use of a simpler model, assuming no-slip conditions in the CCCs, and reducing the number of decision variables necessary for the optimization.
\par

The centroidal dynamics of a system considers the aggregate effects of each rigid link projected onto its center of mass frame, as detailed in \cite{Orin-centroidal}. However, given the low inertia limbs of many quadrupedal platforms and relatively small swinging movements during, this contribution is ignored, and the SRBM is used instead. This simplification significantly simplifies the dynamic constraints on the optimization.
\par

With significant roll or pitch, it is difficult to determine how to prescribe appropriate contact schedules or sequences for landing, especially with the short timeframes high-speed landings would involve.
However, the contact complementarity constraints allow these contact schedules to be directly optimized over without using mixed-integer formulations to characterize making and breaking contact.
They are a set of conditions that require \textit{either} the vertical ground reaction forces \emph{or} the foot's height in the world frame to be zero when the other is positive.
While the CCCs outlined in \cite{Posa-CCC} consider the full Coulomb friction model with sliding, this formulation only considers no-slip constraints, expressed as
\begin{eqnarray}
\phi_i(\mathbf{q}_b, \mathbf{q}_j) \geq 0 \label{eqn:ccc1} \\
\lambda_z \geq 0 \label{eqn:ccc2}  \\
\lambda_z \phi_i(\mathbf{q}_b, \mathbf{q}_j) \leq \epsilon, \label{eqn:ccc3} 
\end{eqnarray}
where $\phi_i(\mathbf{q}_b, \mathbf{q}_j)$ is the signed distance to the ground from the $i^{th}$ foot, $\mathbf{q}_b$ and $\mathbf{q_j}$ are the floating base and joint generalized coordinates respectively, and $\epsilon$ is a slack parameter to encourage convergence. The no-slip condition is enforced with 
\begin{equation}
(\mathbf{r}_{k+1} - \mathbf{r}_k)\lambda_{z,k} = 0, \label{eqn:no-slip} 
\end{equation}
where $k$ is a timestep index in the optimization.
This modification of the CCCs allows for more consistent and faster convergence to solutions, a key requirement for real-time performance.
\par

To further reduce the complexity of the problem, the joint velocities are also excluded as decision variables.
We post-process the optimized joint angle trajectories to find reasonable approximations for the joint velocities.
This likewise improves the convergence and solve times of the NLP.
\par

\begin{figure} 
    \centering
    \includegraphics[width=1.0\columnwidth]{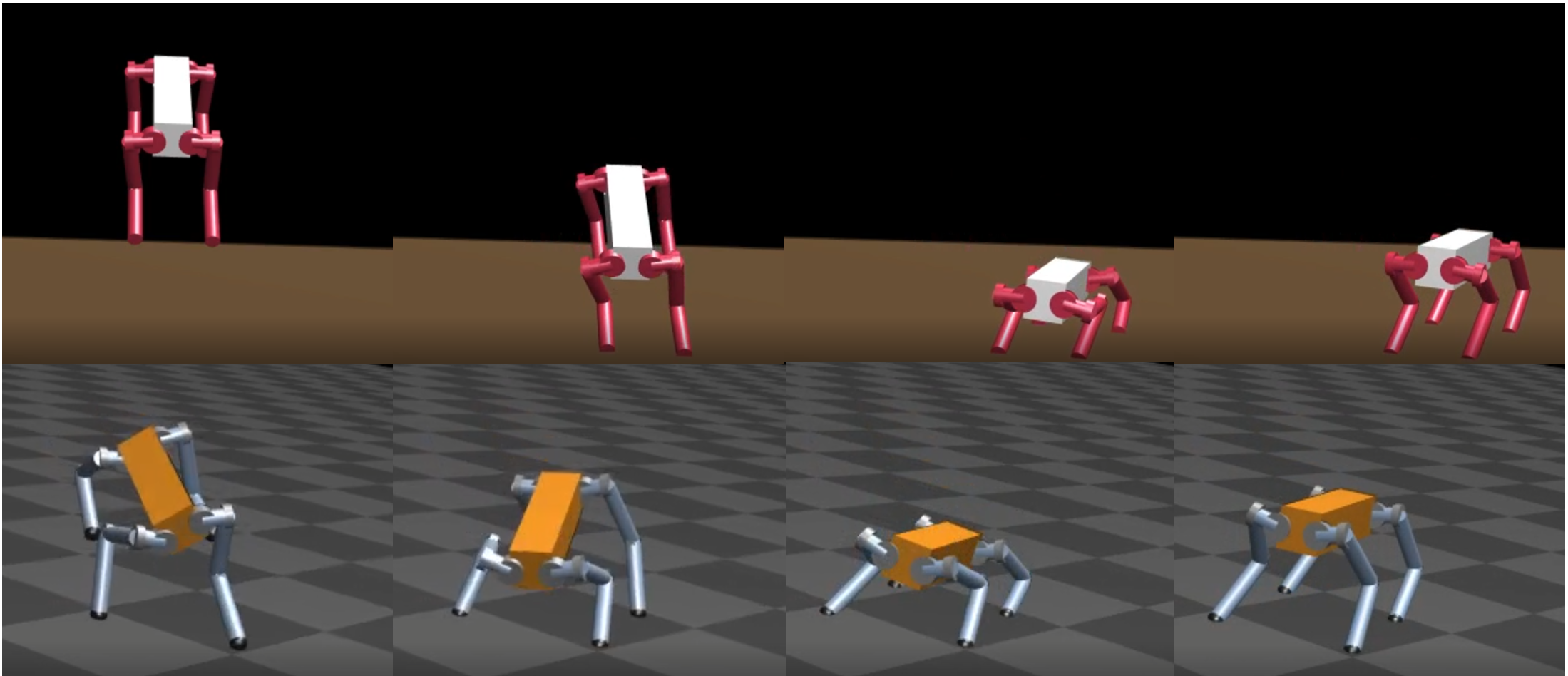}
    \caption{{\bf Landing trajectory optimizations visualized in MATLAB.} The optimization scheme is able to find solutions involving a variety of different contact timings to stabilize the body to a nominal resting height.}
    \label{fig:MATLAB-TO}
\end{figure}

The full optimization formulation can be then be given as
\begin{align}
\min_{\mathbf{x}, \mathbf{u}, \mathbf{q}_j} & \quad ||{(\mathbf{x}_N - \mathbf{x}_{ref})}||_Q^2 \quad \textrm{s.t.} \label{eqn:cost} \\                                    
\textrm{(Dynamics)} & \quad  \dot{\mathbf{x}}_{k+1} = f(\mathbf{x}_k, \mathbf{u}_k), \label{eqn:dynamics} \\                                           
\textrm{(Initial conditions)} & \quad \mathbf{x}(0) = \mathbf{x}_{0}, \\
              & \quad \mathbf{r}(0) = \mathbf{r}_{0}, \\
\textrm{(Contact constraints)} & \quad CCC(\boldsymbol{\lambda}_i, \mathbf{r}_{i,k}) \label{eqn:CCC} \\
              & \quad (\mathbf{r}_{k+1} - \mathbf{r}_k)\lambda_{z,k} = 0,\\ 
              & \quad \lambda_{x,y} \; \epsilon \; \mathcal{F}(\mu, \lambda_z), \label{eqn:friction_cone} \\
\textrm{(Kin. constraints)}              & \quad \mathbf{r}_{i, k}^T\mathbf{r}_{i,k} \leq l_{max}^2, \label{eqn:leg_max}\\
              & \quad \mathbf{r}_{i,k} - g(\mathbf{q}_j) = 0, \label{eqn:leg_kin}\\                                  
\textrm{(Force limits)}
              & \quad |J^T\boldsymbol{\lambda}_k| \leq \boldsymbol{\tau}_{max}, \label{eqn:torque_lim}\\              
\textrm{(Bounds)}
              & \quad \mathbf{r} \; \epsilon \; \mathcal{R}_{x,y,z}(\mathbf{x}_i), \\ 
              & \quad \mathbf{x}_i \; \epsilon \; \mathcal{X}, \\
              & \quad \mathbf{x}(t_f) \; \epsilon \; \mathcal{X}_{term}, \\
              & \quad \mathbf{q}_{j,i} \; \epsilon \; \mathcal{Q}, 
\end{align}
where the decision variables $X = [\mathbf{\Theta}^T \; \mathbf{p}^T \; \boldsymbol{\omega}^T \; \mathbf{\dot{p}}^T]^T \, \in \, \mathbb{R}^{12}$ is the state of the SRBM with roll-pitch-yaw orientation, $U = [\mathbf{r}^T \; \boldsymbol{\lambda}^T]^T \, \in \, \mathbb{R}^{24}$ is the inputs, and $\mathbf{q}_j \, \in \, \mathbb{R}^{12}$ is the joint positions of the legs. The foot Jacobians are represented by $J$, $Q$ is a weighting matrix, $\mu$ is the friction coefficient, $CCC(\boldsymbol{\lambda}_i, \mathbf{r}_{i,k})$ is the set of contact complementarity conditions, and $f(\mathbf{x}_k, \mathbf{u}_k)$, $g(\mathbf{q}_j)$, $\mathcal{F}(\mu, \lambda_z)$, and $\mathcal{R}(\mathbf{x}_i)$ are functions for the dynamics of the SRBM, the forward kinematics of the feet, friction pyramid constraints, and foot kinematic box constraints respectively. The optimization variables are bounded by $l_{max}$, $\boldsymbol{\tau}_{max}$, $\mathcal{X}$, and $\mathcal{Q}$, which are the maximum leg length, torque limits, state limits, and joint limits respectively. The dynamics of the system are propagated through the $N$ timesteps with Euler forward integration. An example of an optimized trajectory in the \texttt{MATLAB} environment can be seen in Figure \ref{fig:MATLAB-TO}. Additionally, instantaneous changes in velocity due to impact are not considered. Due to the light, low-inertia limbs of the Mini Cheetah, it was assumed that impact forces could be ignored at touchdown.

\textbf{Discretization and Initial Conditions: }
To reduce solve times, it was important to keep the number of timesteps $N$ small.
However, certain phases of the trajectory, such as directly after impact, requires high resolution for faithful tracking and smoother force profiles. 
This tradeoff was approached by implementing unevenly discretized timesteps for the phases of the trajectory.
The timesteps nears the final phase of the landing trajectory are larger in comparison to the period directly after impact.
This allows the optimization to find smaller forces over longer time periods to stabilize the robot's state.
\par

The final landing height of the quadruped is unknown, so its initial height is set to "expect" a touchdown immediately at all times.
With constant updates from state estimation, this makes the controller independent of the global height of the robot relative to the ground plane it is landing on.
The initial height of the robot is calculated to be
\begin{align}
    p_{z, h_{min}}      &= \mathrm{min}(S_z{}^BR_0\mathbf{p}_{h,i} \; \forall \; i) \\
    p_{z,0}             &= l_{max} + |p_{z, h_{min}}| + |v_{z, 0}\Delta t_0|, \label{eq:offset-height}
\end{align}
where $^BR_0$ is the rotation matrix from the initial roll-pitch-yaw orientation of the SRBM, $\mathbf{p}_{h,i}$ is the vector from the COM to the $i^{th}$ hip, $S_z$ is a selection matrix to find the z-component of a position vector, $p_{z, h_{min}}$ is the lowest height of the hips of the quadruped, and $\Delta t_0$ is the duration of the first timestep.
While the initial angular velocity of the robot would also contribute to the orientation of the robot at touchdown, small angular velocities are assumed for simplicity. 
\par

\begin{figure} 
    \centering
    \includegraphics[width=1.0\columnwidth]{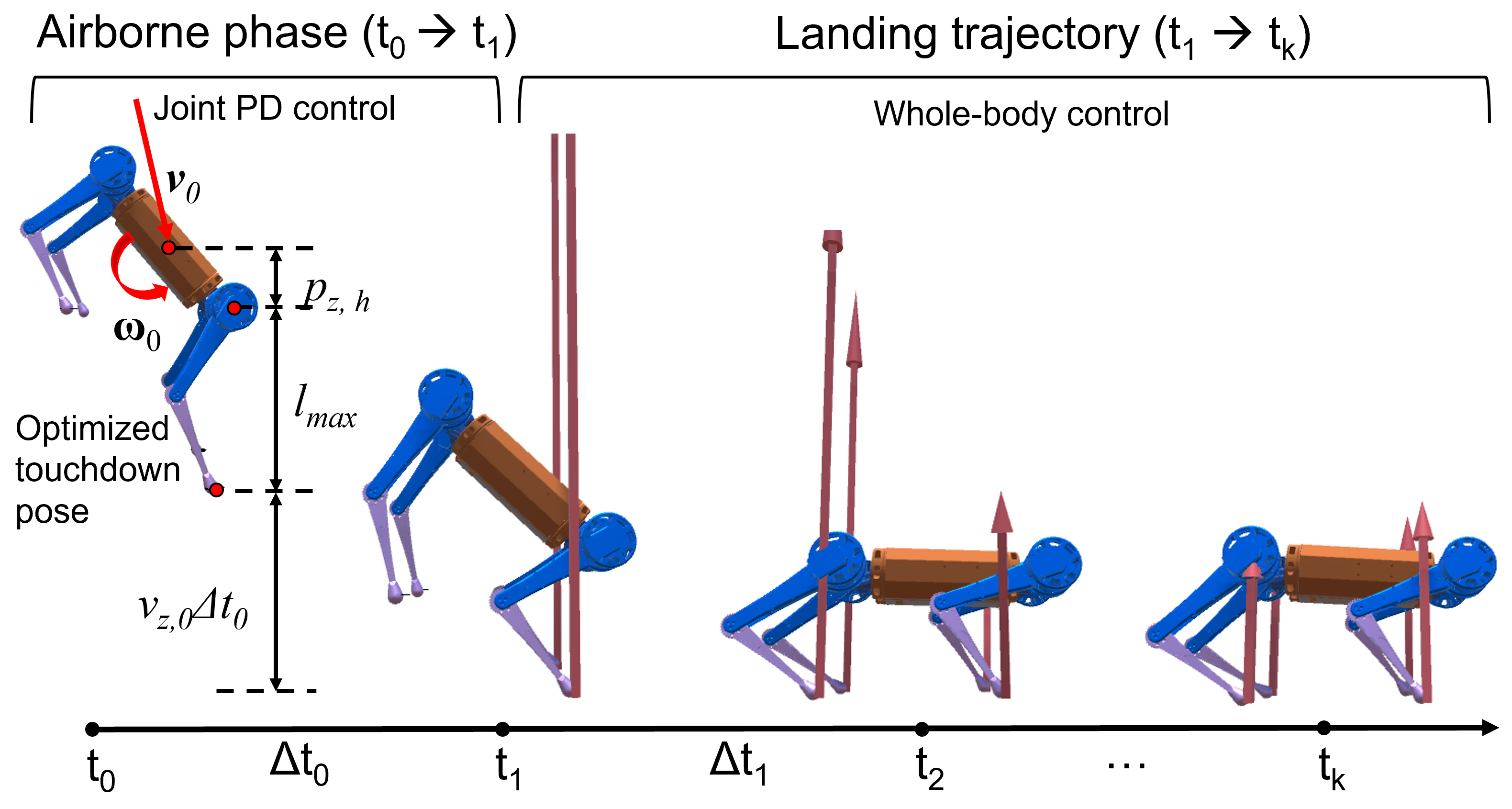}
    \caption{{\bf The initial height setting given by Eq. \ref{eq:offset-height}.} This guarantees that the robot is unable to make contact during the first timestep and to touchdown in the second, allowing for contact locations and poses to be optimized over. If the optimal pose from a previous solution is unavailable, it is initialized with some nominal pose.}
    \label{fig:landing-schematic}
\end{figure}

This offset ensures that the lowest foot is able to reach the ground only \emph{after} the first timestep from the forward Euler integration $p_{z,1} = p_{z,0} + \Delta t_0 v_{z,0}$, as shown in Figure \ref{fig:landing-schematic}. 
While the height of the SRBM could be positioned so that feet begin in contact with the ground as in \cite{Samy-postImpactLanding}, this prevents the NLP from finding optimal footstep locations relative to the body.
With this formulation, the trajectory is always "expecting" to touchdown immediately after the first timestep based on its current velocity, and by solving this problem in a model-predictive fashion, becomes independent of the global coordinates of the body.
\par

\textbf{Cost: } The cost in Eqn \eqref{eqn:cost} consists only of the quadratic penalty between the final state of the optimized trajectory, $X_N$, and a desired final position, $X_{ref}$.
The weighting matrix $Q$ was only non-zero for entries corresponding to velocities, orientation, and a final desired resting height.
It was found that penalizing only the terminal state significantly improved convergence speed, and, in some cases, solution quality.
Introducing running costs on applied forces or states often generated impractical trajectories, where the knees of the Mini Cheetah would penetrate the ground, or significant forces would be commanded before touchdown.
In \cite{Winkler-gaitPhase}, costs similarly seemed to increase solve times and the overall complexity of the NLP.
\par

Without thorough biomechanical study, it is difficult to ascertain which quantities, if any, are best to optimize for landing.
This cost formulation allows for more flexibility in finding landing trajectories instead of imposing costs that may guide the optimization to infeasible solutions.
By only penalizing the final desired state, feasible landing trajectories can be generated quickly and reliably.

\begin{figure} 
    \centering
    \includegraphics[width=0.75\columnwidth]{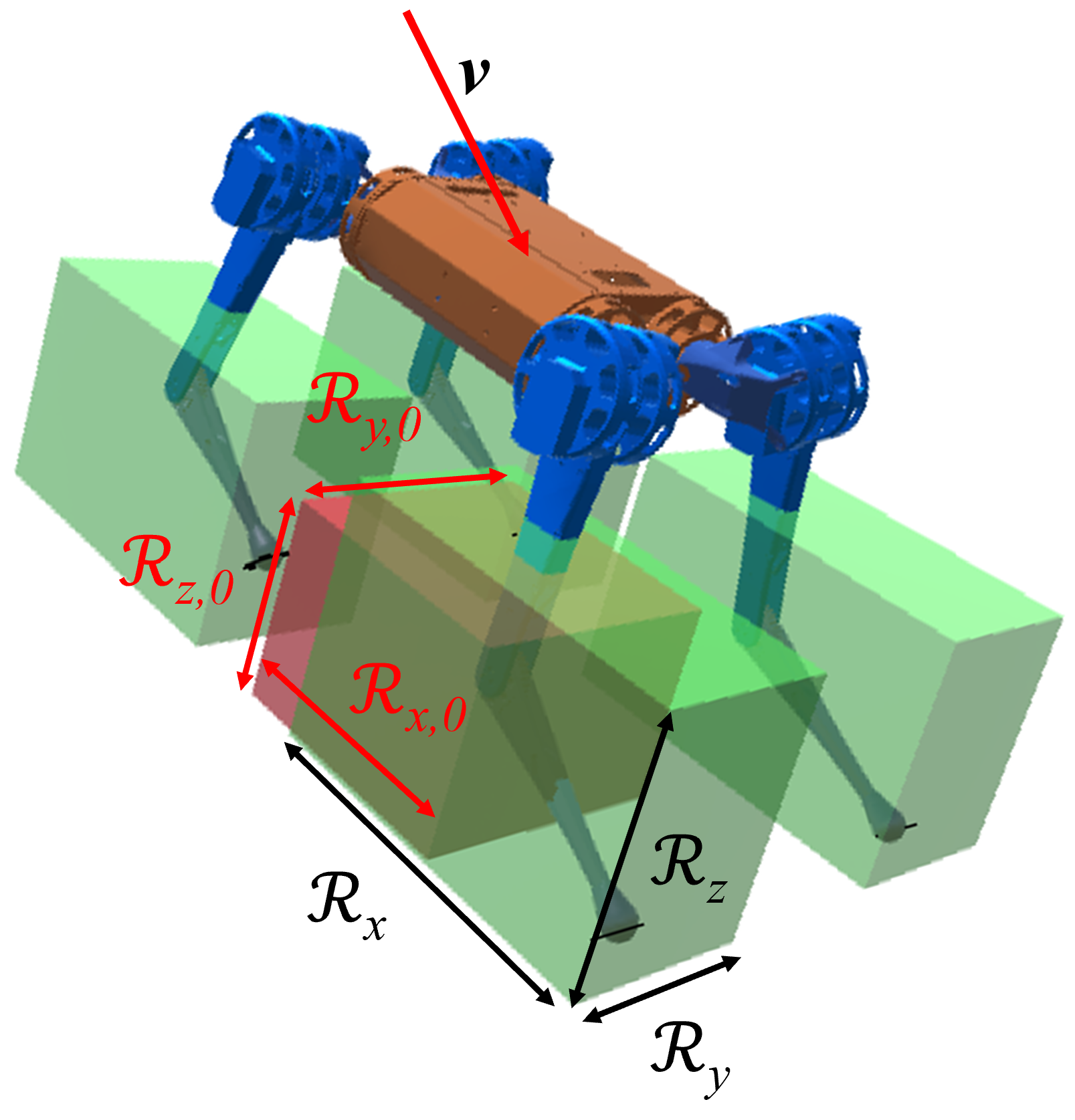} 
    \caption{{\bf Visualized kinematic box limits.} The red box marks the \emph{default} foot position bounds, the dimensions of which are determined by $\mathcal{R}_0$. The kinematic bounds in the optimization (green) adapt to the body velocity $v$ expressed in the body frame and falling conditions as necessary to encourage convergence, the lengths of which are denoted with $\mathcal{R}$.}
    \label{fig:kin-box-limits}
\end{figure}

\textbf{Kinematic Box Limits: }
From initial trajectory optimizations, it was observed that the touchdown angle of the feet correlated strongly with the direction of the COM velocity in the world frame, as shown in Figure [FOOT ANGLE AND VEL PLOT].
Intuitively, it follows that in order to generate force to oppose some velocity, it would be efficient to align the moment arm of the contact point with the direction of that velocity as well.
With this observation, a simple heuristic constraint was added to the optimization that would adjust the kinematic box limits of the quadruped's feet based on its current velocity similar to \cite{Winkler-gaitPhase}, as shown in Figure \ref{fig:kin-box-limits}.
The equations to govern the size of the box limits are given as 
\begin{equation}
    \mathcal{R}_{x,y,z}(\mathbf{x}_i) = \{r|r \leq d|{v \over v_{max}}|\},
\end{equation}
where $d$ is the maximum limit of the box in $x$, $y$, and $z$ directions, $v$ is the $x$, $y$, or $z$ component of the body velocity expressed in the body frame, and $v_{max}$ determines the speed  limit at which the box will extend to the kinematic limits of the leg. 
The values for $d$ and $v_{max}$ were determined experimentally and adjusted based on the solutions of the trajectory optimizations.
By limiting the range of locations the feet could be positioned relative to the body, kinematic limits can be implicitly enforced, improving the convergence of the optimization. 
\subsection{Supervised Learning Framework}
As studied in \cite{Mansard-warmstart}, the initial guess provided to a nonlinear optimization is crucial.
Depending on its quality, the optimization could be attracted to undesirable local minima or fail to find a feasible solution entirely. 
To improve convergence speed as well as solution quality, a neural network is trained in a supervised learning fashion to warm start the nonlinear trajectory optimization outlined in Section III. 
The inputs to the optimization are only the initial orientation, velocity, and angular velocity of the SRBM, which will be denoted $\mathbf{z} = [\mathbf{\Theta}^T \; \boldsymbol{\omega}^T \; \mathbf{\dot{p}}^T]^T \in \mathbb{R}^9$.
Random samples are taken from selected ranges of $\mathbf{z}$ and optimal trajectories are generated for each initial falling condition to train the network. Table 1 details the selected ranges for each element in $\mathbf{z}$. 
\par 

\begin{table}[h]
\caption{Range of sampled input space for supervised learning}
\begin{center}
\begin{tabular}{|c||c|c|c|c|c|c|}
\hline
$\mathbf{z}$ & $\Theta_x$ & $\Theta_y$ & $\Theta_z$ & $\boldsymbol{\omega}$ (r/s) & $v_{x,y}$ (m/s) & $v_z$ (m/s) \\
\hline
Min. & -$\pi/4$ & -$\pi/3$ & -$\pi/2$ & -1 & -1.5 & -6\\ 
\hline
Max. & $\pi/4$ & $\pi/3$ & $\pi/2$ & 1 & 1.5 & -2\\ 
\hline
\end{tabular}
\end{center}
\end{table}
The generated solutions were post-processed so that trajectories with ground penetration or resteps were removed.
The data was normalized with respect to the mean and standard deviation of each variable in the output trajectory.
A neural network with 2 hidden layers with 128 nodes each was trained to generate output trajectories to serve as a warm start to the optimization.

\subsection{Whole-body Controller}
\par

The whole-body controller used in this work is a quadratic program (QP) detailed in \cite{DHKim-highlyDynamic} with a control bandwidth of 500 Hz.
The fast update rates allow for reference trajectories to be tracked closely and stabilized with small deviations in expected states.
An optimized landing trajectory is tracked by the WBC when touchdown is detected. Because timesteps are discretized unevenly, interpolated values of the trajectory are calculated to be tracked by the WBC.

\section{IMPLEMENTATION}
The controller was developed in \texttt{MATLAB} with the \texttt{CasADi} symbolic framework and the \texttt{spatial v2} package from Roy Featherstone \cite{Casadi, spatialv2}.
With \texttt{Casadi}'s internal code generation features, the optimization was exported to \texttt{C++} where the Lightweight Communications and Marshalling package (LCM) was used to communicate between the optimization and the simulation environment and the robot hardware asynchronously \cite{LCM}.
The commercial nonlinear optimization solver \texttt{Artelys Knitro} was used to solve the problem online \cite{KNITRO}.
While the solver \texttt{IPOPT} demonstrated similar solve speeds, \texttt{Knitro} was far more consistent in finding solutions from the range of initial falling conditions given. 
\par

The neural network was developed with \texttt{Pytorch} and similarly exported to \texttt{C++} with the \texttt{LibTorch} packages.
Without the overhead of the \texttt{MATLAB} interface, the optimizations were able to be solved faster as well. 

The optimization is run as a process separate from the main body of the software on the Mini Cheetah.
Requests for optimizations are sent via LCM from the robot's state estimator and received when a solution is found.
If a solution is not found within 300 ms, a new request is made and the optimization is restarted.
Decoupling the optimization from the main processes on Mini Cheetah in this manner allows for asynchronous, model-predictive control.

Touchdown is detected when a threshold on the joint velocities of the legs is exceeded.
The latest feasible trajectory is then interpolated and tracked with the QP-based whole-body controller from \cite{DHKim-highlyDynamic}.
While there have been learned and probabilistic approaches to detecting contact proprioceptively \cite{Bledt-contact, Lin-MLcontact}, joint encoder feedback is used instead.
Due to its low response time of around 3 ms and the impulsive nature of landing that could be difficult to learn, encoder data was used as a touchdown trigger for its simplicity.

\section{RESULTS}
\subsection{Trained Trajectory Generation}
The network was trained with roughly 1500 samples of trajectory data over a range of falling conditions. The state space was uniformly sampled over, but the majority of the successful landings tended to be near the horizontal body posture, as landing in a near-horizontal position is the simplest landing to stabilize.
Samples from the space of successful landing conditions are shown in Figure \ref{fig:nn-solve-time}. 

\subsection{Real-time Performance}
\begin{figure} 
    \centering
    \includegraphics[width=1\columnwidth]{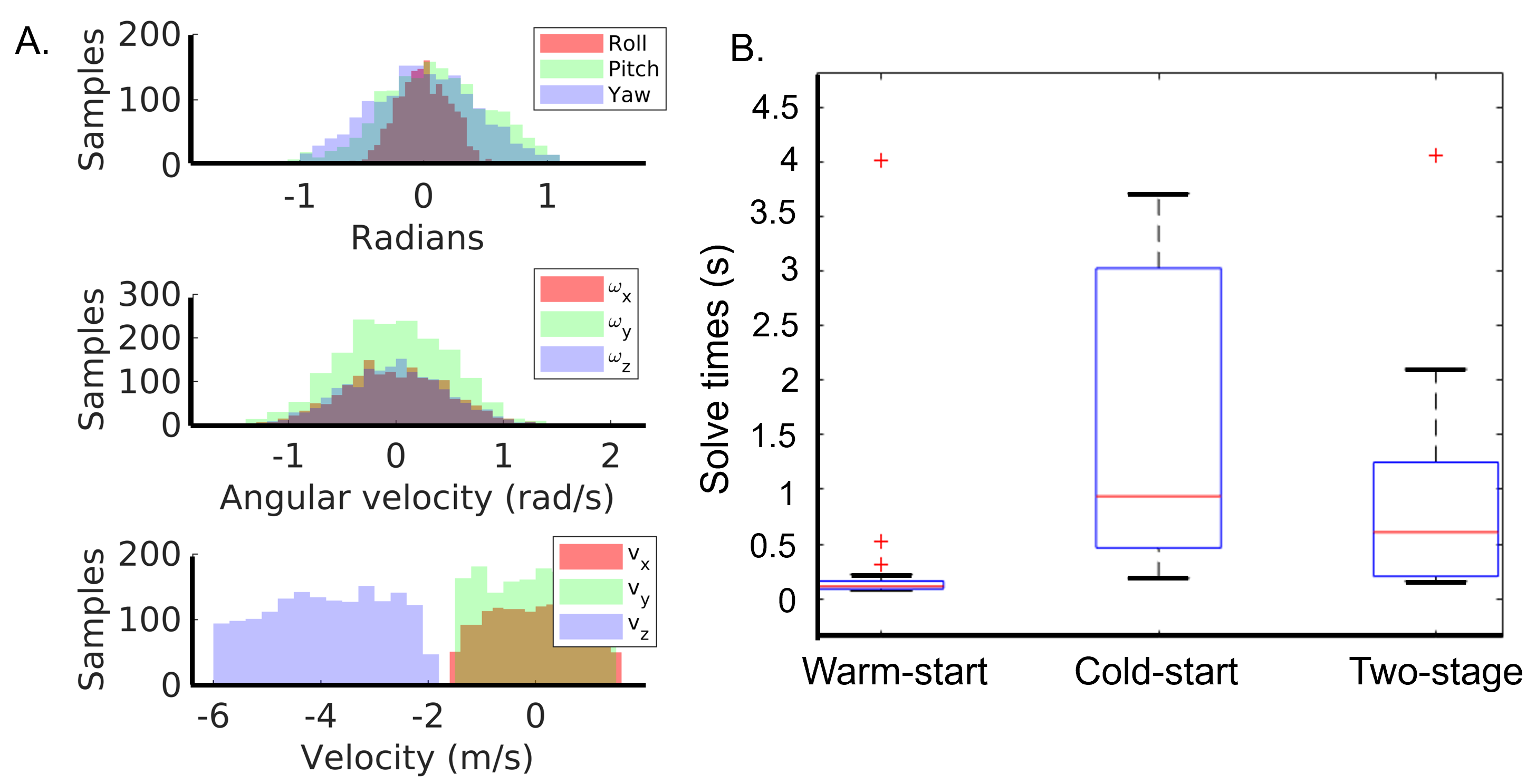} 
    \caption{{\bf Training data sampling and warm-start evaluation.} A: sampled space of falling conditions for training data. B: comparisons of solve times for neural net warm-started optimization, cold-started optimization, and two-stage warm-started optimization, from left to right.}
    \label{fig:nn-solve-time}
\end{figure}

By warm starting the optimization with the trajectories generated by the neural net, there is a significant decrease in computation time that allows for real-time performance to be possible.
On a typical desktop computer in the \texttt{MATLAB} environment, $\mathbf{z}$ was randomly sampled across the same ranges as in Table 1 and optimized with the warm-start from the neural net.
This was compared with the solve times from cold-starts and a two-stage optimization process.
The two-stage optimization involved solving the NLP outlined in Section III \emph{without} kinematic decision variables, and using its solution as a warm start to the full kino-dynamic optimization.
As shown in Figure \ref{fig:nn-solve-time}, the neural-net warm started optimization showed speedups of roughly ~5-10 times over the range of falling conditions.
Even with the overhead of \texttt{MATLAB}, solve times were in the range of 5-8 Hz, allowing for real-time performance onboard the Mini Cheetah.

\subsection{Simulation}
The Mini Cheetah was initialized in various falling configurations (orientation and velocities) in the Robot-Software environment and simulated to test recovery behavior.
The desired final state for $X_{ref} = [\mathbf{\Theta}^T \; \mathbf{p}^T \; \boldsymbol{\omega}^T \; \mathbf{\dot{p}}^T]^T$ was set to be all zero except the height, which was set to 0.25 m.
Zero entries in the weighting matrix $Q$, ensures no penalty was applied on the final yaw, $p_x$, or $p_y$.
\par

\begin{figure} 
    \centering
    \includegraphics[width=1\columnwidth]{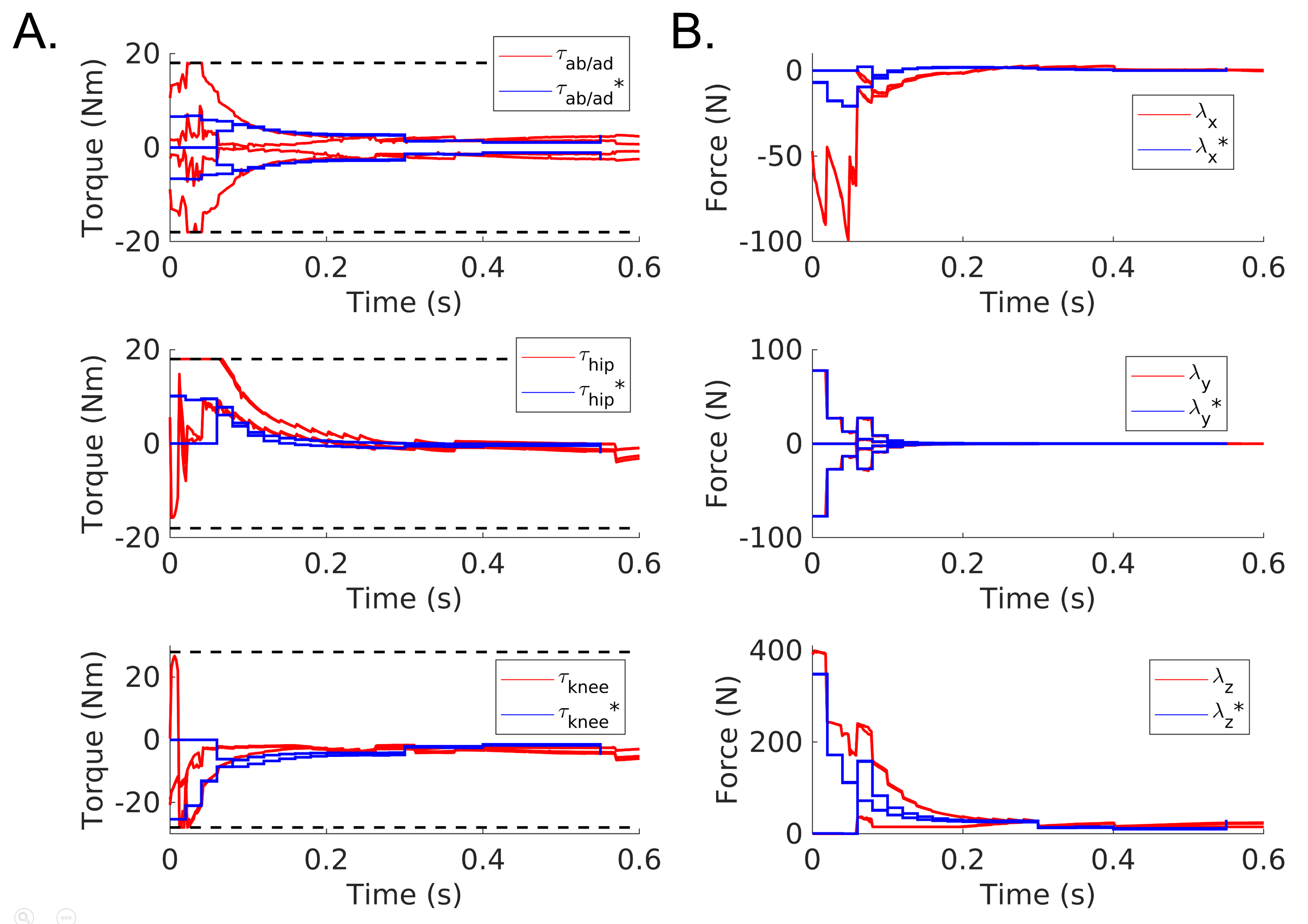} 
    \caption{{\bf Actuation limits during simulated landing.} A: desired ($\tau*$) and actual ($\tau$) torques from the simulation environment. The black dotted lines represent the absolute torque limits of the individual motors. B: desired ($\lambda*$) and actual ($\lambda$) ground reaction forces commanded by the whole-body controller. The Mini Cheetah is able to stabilize itself from significant drops while respecting the torque limits of the system.}
    \label{fig:sim-plots}
\end{figure}

The Robot-Software simulation evaluates its dynamics at 1000 Hz and includes motor constraints, rotor inertias, and impact calculations in its environment.
A friction coefficient of $\mu = 0.75$ was used, and it was assumed that there was no state estimation error in the simulation.
Torque and force plots for a landing trajectory from a 2.5 m high pitched landing with small lateral velocities are shown in Figure \ref{fig:sim-plots}.
While the impact on touchdown causes a high initial tracking error, this quickly goes to zero within the first 0.1 s, and both torque and force profiles are tracked closely.
With the adjustments from the whole-body controller, the trajectories are well followed even with significant disturbances. 
Deliberate errors in velocity, orientation, or angular velocity were used as inputs to the optimization, but the controller was often able to stabilize to a nominal resting position with small errors in the expected falling conditions $\mathbf{z}$. The authors plan to more fully characterize the robustness of this controller in future work.
\par

To test the limits of the hardware in simulation, the Mini Cheetah was dropped horizontally from a height of 8 m, corresponding to vertical velocities of roughly -12 m s$^{-1}$. The controller was able to stabilize the fall, but further work must be done to investigate the effect of the impacts and post-impact velocities at these higher speeds. 

Recovering from landing is far more sensitive to roll and lateral velocity than pitch and longitudinal velocity.
This is intuitive given the design of the Mini Cheetah, where only the single ab/ad motor is dedicated to stabilizing significant disturbances in body roll and lateral velocity.

\subsection{Hardware}
To verify the controller on hardware and ensure safe landings, it was critical that the state estimation of the Mini Cheetah was as accurate as possible. 
However, there was severe and constant drift in state estimates once the robot is in the air because kinematics based estimation is not possible while airborne. With unreliable state estimation, falling conditions were instead hard coded to be approximately similar to the state of the robot at touchdown instead.
\par

\begin{figure} 
    \centering
    \includegraphics[width=.6\columnwidth]{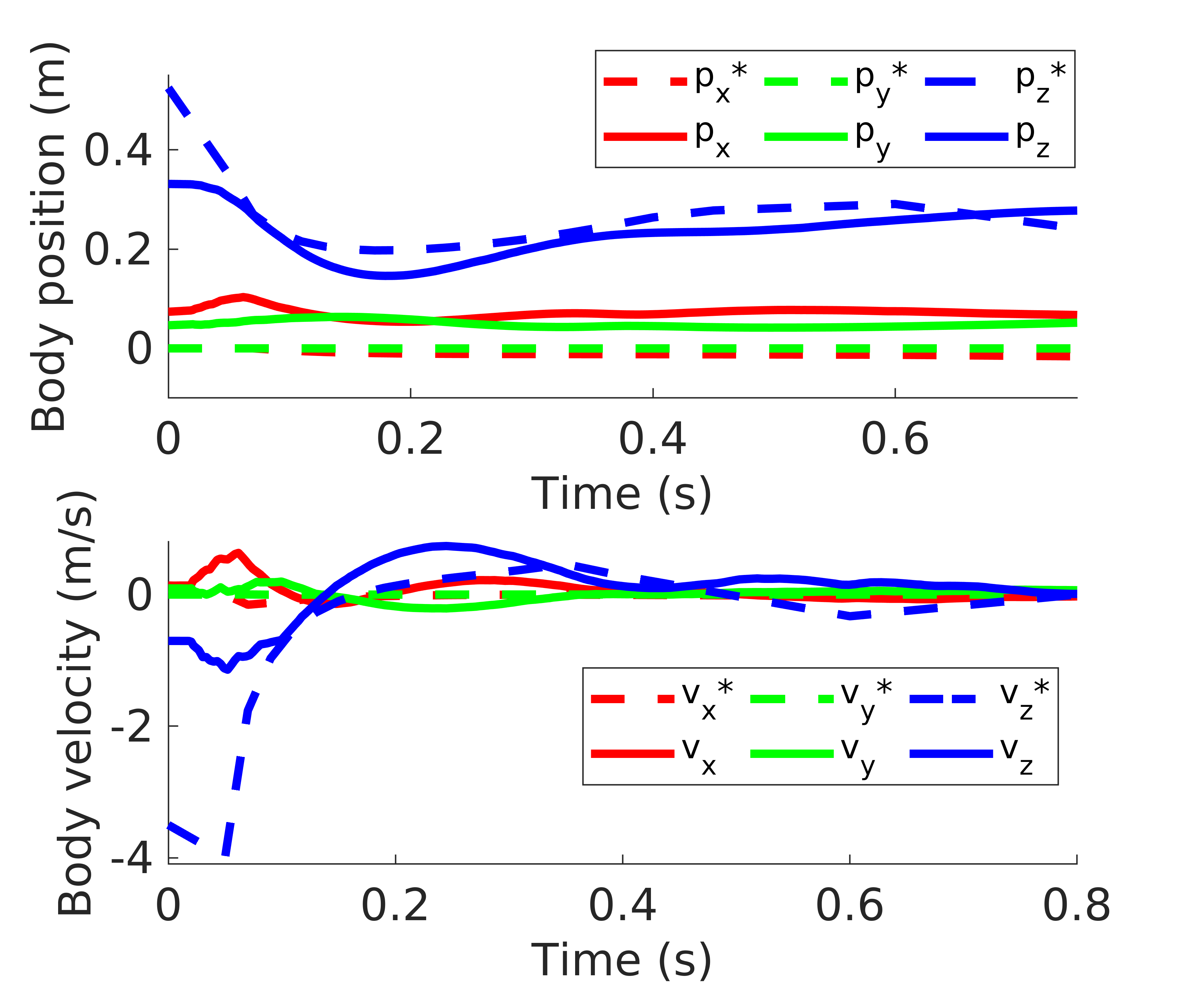} 
    \caption{{\bf Trajectory tracking on hardware.} Desired body posture ($p*$, $v*$) and body posture from state estimation ($p$, $v$). Note that the velocity discrepancy is due to pausing the state estimation until touchdown.}
    \label{fig:hardware-plots}
\end{figure}

Several horizontal drops were performed from heights of up to 2 \si{\meter} successfully, as shown in Figure \ref{fig:mc-landing}. 
Although hard-coded conditions were used as inputs to the optimization, the Mini Cheetah was released by hand, causing deviations from the expected falling state. 
Despite these errors, the Mini Cheetah was able to recover to a nominal resting height, as shown in the plots in Figures \ref{fig:hardware-plots}. The state estimator was paused in a resting state until touchdown was detected because of its drift, causing a significant difference between the estimated and actual velocities of the quadruped at touchdown, but the WBC is able to stabilize the robot about the planned trajectory. 
\par

Because of the unreliable state estimation, drops with significant pitch, roll, or lateral velocities were not tested to prevent damage to the robot.
While simulation results suggest these landings would also be possible, we found that significant orientation errors in the state estimator quickly caused the controller to become unstable in the Robot-Software environment. 
\par

\section{CONCLUSIONS}
This paper presents a control framework that is able to reason about optimal contact locations and timings in real-time, and demonstrates successful landings in both simulation and hardware. By modifying the contact constraints and model considered, dynamic trajectories can be planned independently from prescribed contact modes at real-time rates. 
\par

Future work will involve hardware upgrades to the Mini Cheetah, implementing onboard vision, and characterizing the robustness of the controller.
Processor and power board upgrades to the Mini Cheetah will enable it to solve the presented formulation completely untethered and onboard, and with greater margin for motor failure.
The authors also plan to integrate an event camera into the Mini Cheetah for more accurate localization and state estimation.
Additionally, while it was shown that the controller was able to stabilize itself from small errors in the falling conditions, we have yet to characterize this.
In the future, the robustness of the controller and its performance over uneven terrain will be explored in greater detail.




\section*{ACKNOWLEDGMENT}
This work was supported by Naver Labs and the Centers for ME Research and Education at MIT and SUSTech. The authors would like to thank Elijah Stanger-Jones and Aditya Mehrotra for their assistance in hardware experimentation, and the other members of the Biomimetic Robotics Lab for their insight and support in this work. 
\addtolength{\textheight}{0cm}   

\printbibliography


\end{document}